\documentclass[10pt,twocolumn,letterpaper]{article}

\usepackage{iccv}
\usepackage{times}
\usepackage{epsfig}
\usepackage{graphicx}
\usepackage{amsmath}
\usepackage{amssymb}
\usepackage{multirow}
\usepackage{graphicx}
\usepackage{subcaption}
\usepackage{bm}
\usepackage{soul}
\usepackage{hhline}
\usepackage{adjustbox}
\usepackage{caption}

\usepackage[pagebackref=true,breaklinks=true,letterpaper=true,colorlinks,bookmarks=false]{hyperref}

 \iccvfinalcopy 

\def\iccvPaperID{1070} 

\ificcvfinal\fi
\begin{document}

\title{Improving Classification by Improving Labelling: \\Introducing Probabilistic Multi-Label Object Interaction Recognition}

\author{Michael Wray \qquad Davide Moltisanti \qquad Walterio Mayol-Cuevas \qquad Dima Damen\\
University of Bristol\\
{\tt\small <FirstName>.<LastName>@bristol.ac.uk}
}

\maketitle

\begin{abstract}
   
   
This work deviates from easy-to-define class boundaries for object interactions. For the task of object interaction recognition, often captured using an egocentric view, we show that semantic ambiguities in verbs and recognising sub-interactions 
along with concurrent interactions result in legitimate class overlaps (Figure~\ref{fig:magic_example}). We thus aim to model the mapping between observations and interaction classes, as well as class overlaps, towards a probabilistic multi-label classifier that emulates human annotators.
   
Given a video segment containing an object interaction, we model the probability for a verb, out of a list of possible verbs, to be used to annotate that interaction. The probability is learnt from crowdsourced annotations, and is tested on two public datasets, comprising $1405$ video sequences for which we provide annotations on $90$ verbs. 
We outperform conventional single-label classification by 11\% and 6\% on the two datasets respectively, and show that learning from annotation probabilities outperforms majority voting and enables discovery of co-occurring labels.
\end{abstract}

\vspace*{-10pt}
\section{Introduction}
\label{sec:intro}


Previous works on object interaction recognition, captured using egocentric cameras,
use semantically distinct verbs as class labels.
Labels are chosen by a majority vote~\cite{Motwani12} or by specifically selecting verbs 
that make annotating unambiguous~\cite{de2008guide, Fathi2012, Pirsiavash12,  yumi2014first}.
In this work, we argue that recognising object interactions as a standard one-vs-all classification problem stems from three incorrect assumptions (IAs), see Figure~\ref{fig:magic_example}:

\begin{figure}[t]
\begin{center}
   \includegraphics[width=\linewidth]{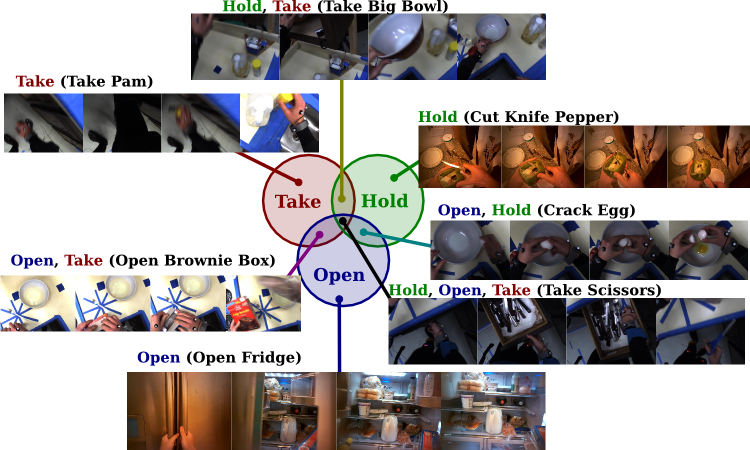}
\end{center}
   \caption{For three annotation verbs \textit{Take, Hold, Open}, and seven object interactions, the class boundaries overlap. For the act of `Taking a Bowl' (top), human annotators would use \textit{hold} and \textit{take} as valid labels, but not \textit{open}. Single-label one-vs-all classifiers cannot capture these class overlaps.}

\label{fig:magic_example}

\end{figure}

\noindent \textit{\underline{IA1:} object interaction classes can be self-contained and have strict boundaries}. As the number of verbs increases, classes overlap.
For instance take three verbs: \textit{open}, \textit{switch~on} and \textit{turn~on}. The act of \textit{turning~on} a tap could be referred to as \textit{opening} the tap but not as \textit{switching} it \textit{on}. On the contrary, \textit{turning~on} the hob is \textit{switching} it \textit{on}.
A classifier would struggle to define boundaries between these verbs when no hard boundary may exist semantically.

\noindent \textit{\underline{IA2:} a sequence can be split into segments, each with exactly one object interaction}.
Temporal granularities are difficult to define in practice. A video could show someone \textit{holding} and \textit{filling} a cup whereas another could include the user \textit{holding} a knife to \textit{cut}, and a third \textit{holding} and \textit{cracking} an egg.
Frequently, multiple object interactions are performed simultaneously like \textit{turning on} a tap to \textit{fill} a cup.

\noindent \textit{\underline{IA3:} it is a binary decision whether a verb could be used to label an object interaction.} Given crowdsourced annotations, we show that some verbs are more likely to be chosen than others to annotate object interactions.
Human annotators are more likely to use the verb \textit{open} when referring to \textit{pulling} a fridge door open. A method that can learn the probability of using a verb to annotate a video would better emulate a human annotator.


We present the following contributions in this paper:

\vspace*{-8pt}
\begin{itemize}
\item We employ crowdsourcing to annotate egocentric videos using a list of $90$ verbs. We collect annotations that offer meaningful statistics and informative semantics on multi-label recognition for two public datasets.
\vspace*{-3pt}
\item We reformulate object interaction recognition as a probabilistic multi-label classifier, trained using a two-stream CNN. We show that this model can better capture the relationships between observations and annotations, as well as the co-occurrences of labels.
\vspace*{-3pt}
\item We 
demonstrate that using a multi-label probabilistic classifier, as opposed to a single label chosen from majority vote, provides higher recognition accuracy.
\end{itemize}
\vspace*{-4pt}

The rest of the paper is as follows: A review of recent and relevant works is contained in Section~\ref{sec:related_work} . 
Details into how we collected the annotations for the two datasets can be read in Section~\ref{sec:annotations}.
We then formulate the method in Section~\ref{sec:method} and Section~\ref{sec:experiments} contains the experiments and results. 
\vspace*{-3pt}


\section{Related Work}
\label{sec:related_work}

\noindent \textbf{Action Recognition} \hspace{3pt}
Action Recognition has largely focussed on a single label classification approach. Hand crafted features dominated most seminal action recognition works ranging from those that have used spatio-temporal interest points~\cite{scovanner20073, laptev2008learning, willems2008efficient, klaser2008spatio} with a bag of word representation to trajectory-based methods~\cite{Wang2011, Wang2013}, encoded using Fisher Vectors~\cite{perronnin2010improving}. Features were typically classified using one-vs-all SVMs.
Within the egocentric domain other features such as gaze~\cite{Fathi2012}, hand~\cite{ishihararecognizing,kumarfly} or object specific features~\cite{fathi2013modeling, Damen2014a, McCandless13,ren11,Pirsiavash12} were also incorporated.

More recently, CNNs have been trained end-to-end to determine actions for traditional~\cite{ji20133d, karpathy2014large, tran2015learning, lea2016segmental, feichtenhofer2016convolutional} and egocentric~\cite{poleg2016compact, zhou2016cascaded, Ma16} datasets.
These approaches have consistently outperformed hand crafted features for recognising actions and use both spatial information and temporal information via optical flow.
For instance, Ma~\etal~\cite{Ma16} use an appearance-based CNN to localise objects and a temporal CNN to detect actions. These are then fused to detect the activity. All previous approaches on object interaction recognition, though, define an object interaction class label as a verb-noun combination, attempting to distinguish `take cup' from `take knife'. This makes them less generalisable to the same interactions but with novel objects. In this work, we attempt to label object interactions using \textit{verb} labels solely in order to study class boundaries, \ie~\textit{what is the space of interactions that could be labelled with `take'?}. 


\noindent \textbf{Semantics and Multiple Labels} \hspace{3pt}
Predicting multiple labels for the same input has not been attempted for action recognition, but is increasingly popular for 
predicting multiple objects present in the scene (\eg predicting the presence of a person, a TV and a table in the same image).
These works are motivated by the variety of appearances of objects within the scene, which makes global CNN features unable to correctly predict multiple labels.
Approaches thus use object proposals ~\cite{gong2013deep, yang2016exploit} or learn spatial co-occurrences of labels using RNNs/LSTMs~\cite{zhang2016multi, wang2016cnn}. 


We note that our work differs from other multiple label classifiers in that previous works are focussed on predicting multiple, unrelated -- though co-occuring -- labels, \eg human and table, whereas we are also interested in predicting annotations that can be used to describe the same action, such as \textit{place} and \textit{put}.

There is little related work in action recognition that deals with semantics or differing verbs.
Motwani and Mooney~\cite{Motwani12} collect sentences for action videos. Their method chooses the most frequent verb out of the annotated sequences which is then, using {W}ord{N}et~\cite{miller1995wordnet}, clustered into activity clusters.
Wray~\etal~\cite{wray2016sembed} showed that when annotators were given freedom of choice over which verbs to label, a variety of verbs were chosen, such as \textit{kick}, \textit{pedal} and \textit{fumble}.
In most cases, a small number of common verbs were chosen showing a long tail distribution for verbs.
In their work, videos are labelled with a single verb but they use semantic knowledge in an attempt to relate these verbs.
Another work to use free annotations is Alayrac~\etal~\cite{Alayrac15learning}. The authors automatically use narrations in 
{Y}ou{T}ube videos 
to align 
and cluster action sequences of the same task.
A new activity dataset was collected by Sigurdsson~\etal~\cite{sigurdsson2016hollywood} containing videos acted out using scripts which were created by users from a list of verbs and nouns.
The authors use a diverse set of actions and vocabulary to describe their dataset. 

\noindent \textbf{Caption Generation} \hspace{3pt}
Semantics and annotations have also been studied by the recent surge in generating captions for images~\cite{mao2014deep, devlin2015language, anne2016deep} and video~\cite{venugopalan2014translating, yao2015describing, yu2016video}. These works assess success by the acceptability of the generated caption, and do not focus on generating multiple valid captions per object or action. For instance, Yu~\etal~\cite{yu2016video} use a hierarchical RNN to generate paragraphs of captions for longer activity videos by feeding sentences through a second RNN. 

In summary, while many works have attempted similar problems, none have focussed on multi-label classification for action recognition. 
We are particularly interested in classification, i.e. predicting whether a label could be used to describe an input video, yet wish to accommodate semantically related labels, or co-occurring actions common in daily object interactions. We aim for an approach that extends beyond the handful list of verbs commonly used in action recognition works.
\vspace*{-6pt}
\section{Annotations}
\label{sec:annotations}
\vspace*{-5pt}
As all previous works label an action or interaction sequence with a single label/verb, no previously published datasets are suitable for training/testing probabilistic or multi-label classification of object interactions.
Instead of collecting a new dataset, we provide annotations for the two largest egocentric datasets for object interactions: CMU~\cite{de2008guide} and GTEA+~\cite{Fathi2012}
In this Section, we describe the procedure followed to collect annotations (\ref{subsec:ann_procedure}), as well as the statistics of the collected annotations (\ref{subsec:ann_statistics}).

\subsection{Annotation Procedure}
\label{subsec:ann_procedure}
\begin{figure*}[t]
\centering
  \makebox[\linewidth][c]{%
  \begin{subfigure}[b]{0.535\textwidth}
    \centering
    \includegraphics[width=\textwidth]{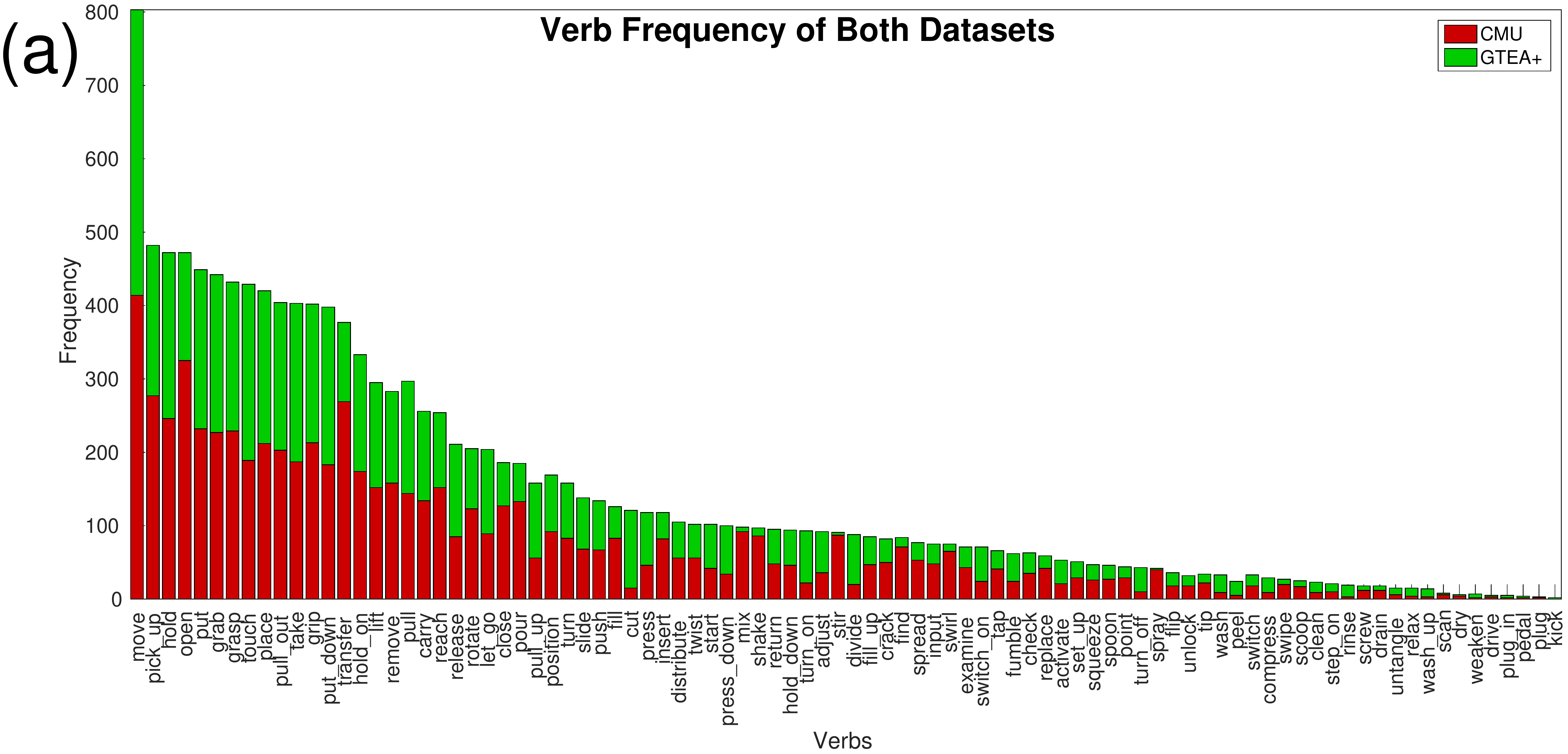}
    \label{fig:verb_freq}
  \end{subfigure}
  \hfill
  \begin{subfigure}[b]{0.535\textwidth}
    \centering 
        \includegraphics[width=\textwidth]{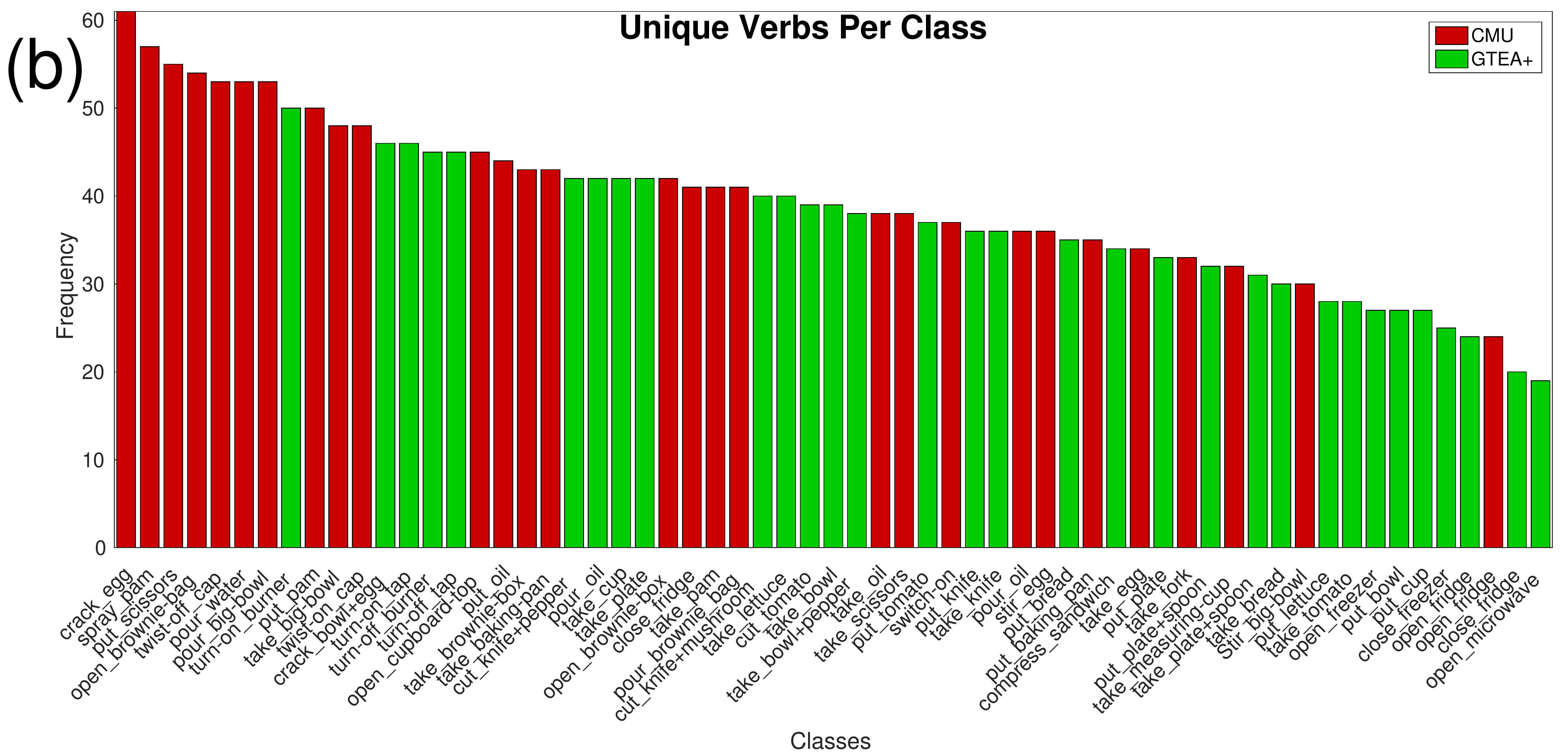}
    \label{fig:annotation_unique}

  \end{subfigure}
  }
  \\[-3ex] 
  \makebox[\linewidth][c]{%
  \begin{subfigure}[b]{0.53\textwidth}
    \centering 
        \includegraphics[width=\textwidth]{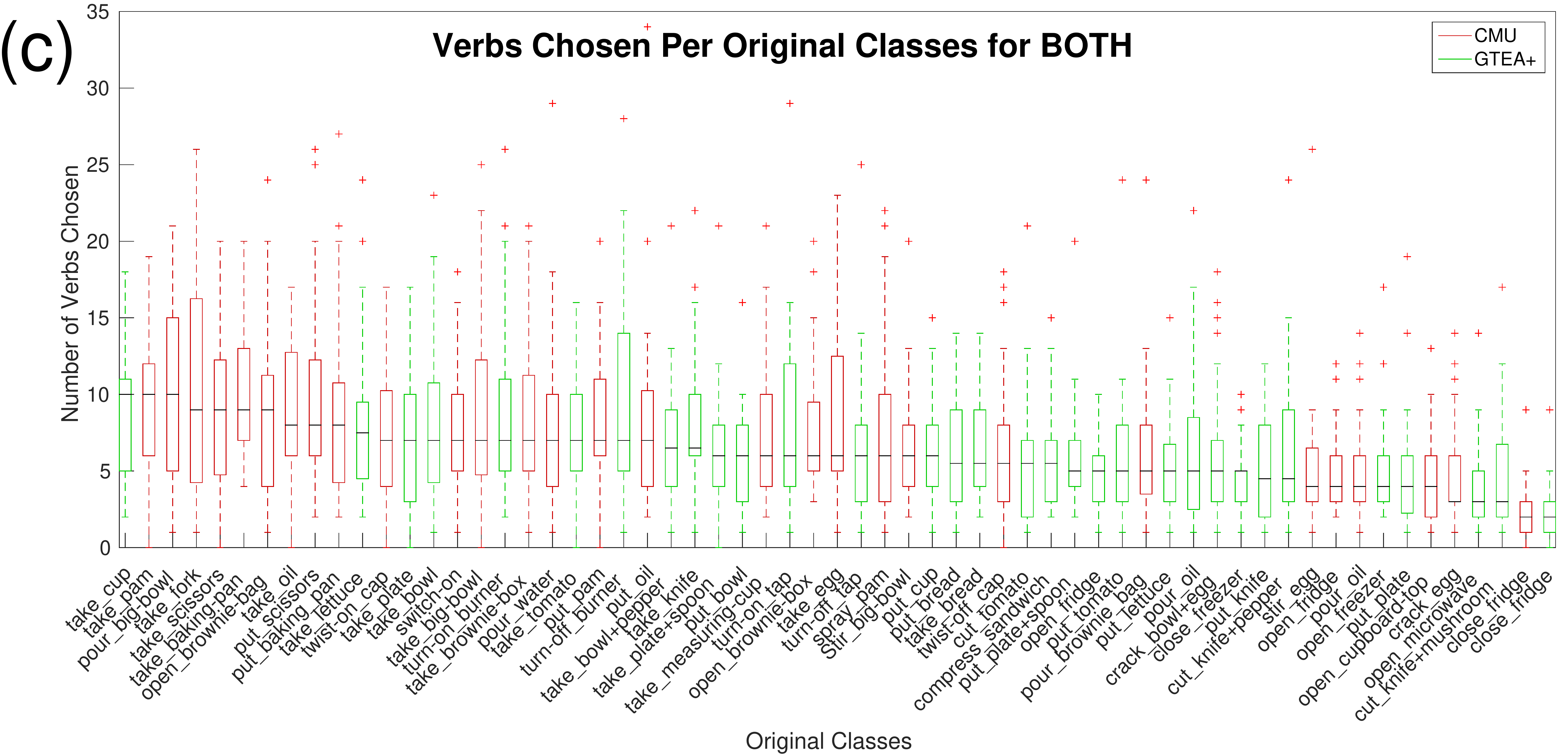}
    \label{fig:annotation_chosen}
  \end{subfigure}
  \hfill
  \begin{subfigure}[b]{0.53\textwidth}
    \centering 
    \includegraphics[width=\textwidth]{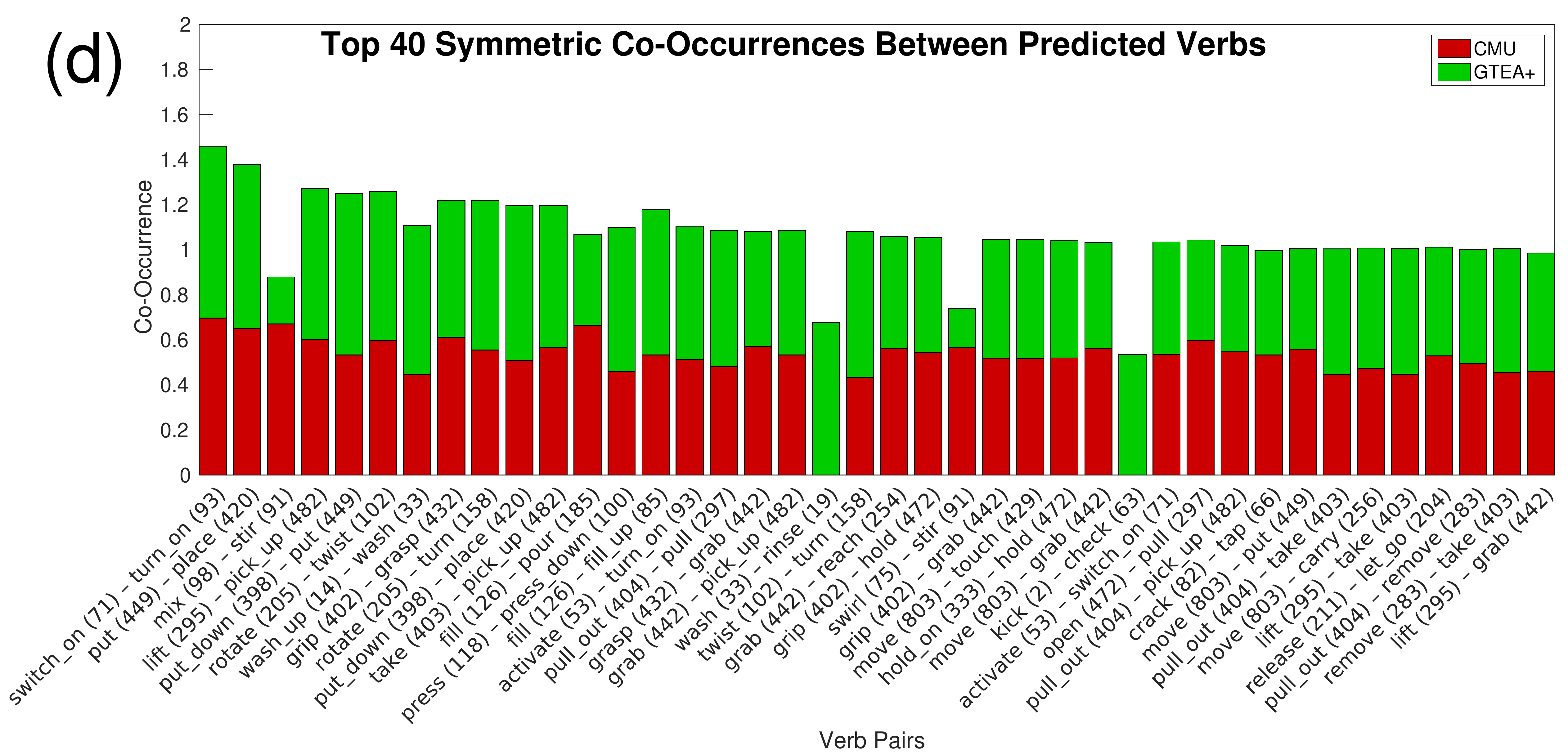}
    \label{fig:annotation_symm}
  \end{subfigure}
  }
  \caption{\textbf{Annotation Statistics.} Figure best viewed in colour - CMU (red), GTEA+ (green) (a) Number of annotations per verb. (b) Number of unique annotations per class. (c) Box plot for the number of verbs chosen per annotator for each class. 
  (d) Top 40 symmetric co-occurrences of verb pairs ordered for both datasets and displayed to show the contribution of each dataset separately.}
  
  \label{fig:annotation}
\end{figure*}

We accumulated a list of potential verbs for labelling by merging all unique verbs or phrasal verbs in the ground-truth annotations of three egocentric datasets~\cite{de2008guide, Fathi2012, wray2016sembed}.
The three datasets have a total of $427$ verb-noun classes of which there are  $93$ unique verbs. For example: `Spray Pam' from~\cite{de2008guide} contributed \textit{spray}, `Compress Sandwich' from~\cite{Fathi2012} gave \textit{compress} and `Touch Card' from~\cite{wray2016sembed} added \textit{touch}.
We found some overlaps between datasets, a total of $23$ verbs are present in at least two datasets, with generic verbs such as \textit{open}, \textit{put} and \textit{take} being present in all three.
Due to the kitchen activities taking place in~\cite{de2008guide, Fathi2012}, these have common verbs like \textit{crack}.
Additionally,~\cite{wray2016sembed} introduced 62 new verbs, such as \textit{pedal}, \textit{scan} and \textit{screw}, as the annotators were given free verb choice and the activities in this dataset extend beyond the kitchen to include using gym equipment.
From the list of $93$ verbs, we removed $3$ verbs; \{\textit{read, rest, walk}\} as we focused solely on object interactions, resulting in the $90$ verbs that form the basis for all collected annotations (see Figure~\ref{fig:annotation}(a) for a complete list).

We annotated both datasets, CMU~\cite{de2008guide} and GTEA+~\cite{Fathi2012}, by selecting a reference video from each class. The reference video was chosen such that the ground truth action was clearly visible with good lighting in an attempt to reduce annotation noise.
We instructed Amazon Mechanical Turk (AMT) workers to choose all verbs out of the list of $90$ verbs that were correct labels for the reference video.
Each video was annotated by a minimum of $30$ and a maximum of $50$ workers to give a distribution for each of the videos.
In total, we collected annotations from
$1933$ AMT workers.

\begin{figure}[t]
\centering
  \begin{subfigure}[b]{0.48\textwidth}
    \centering 
    \includegraphics[width=\textwidth]{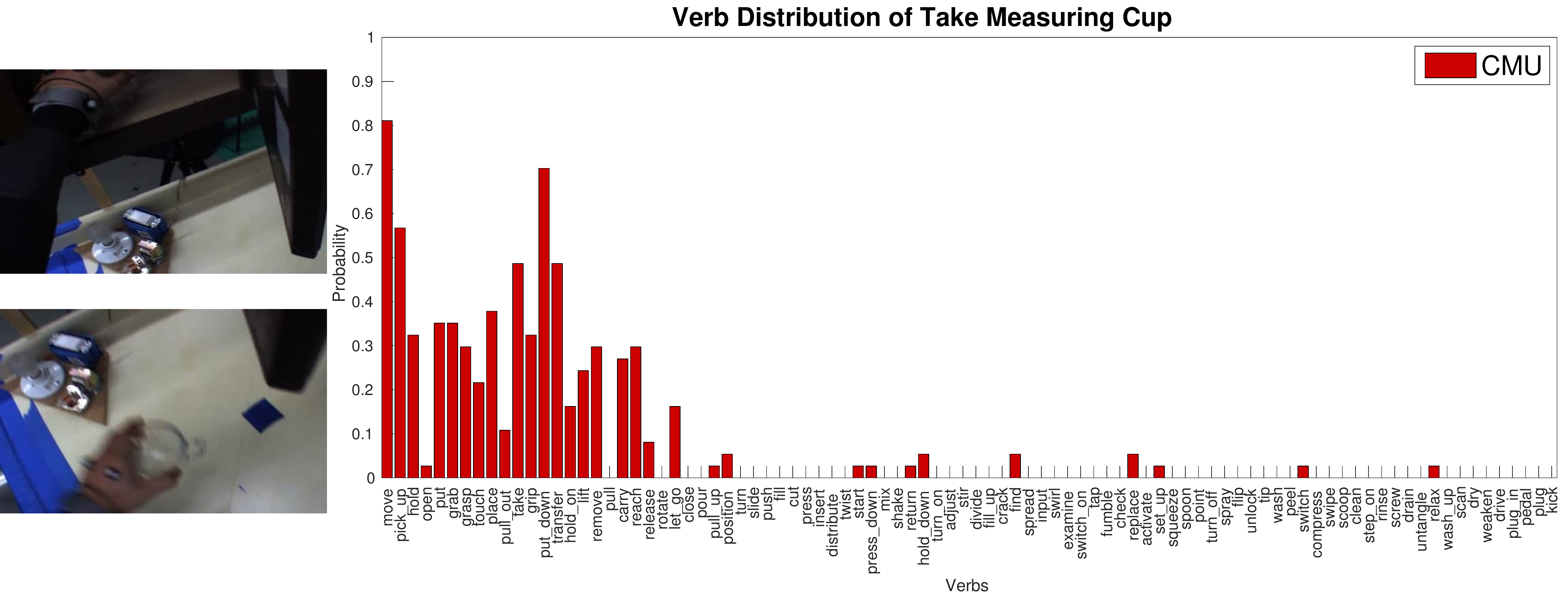}
    \label{fig:crack_egg}
  \end{subfigure}
  \\[-3ex]
  \begin{subfigure}[b]{0.48\textwidth}
    \centering 
    \includegraphics[width=\textwidth]{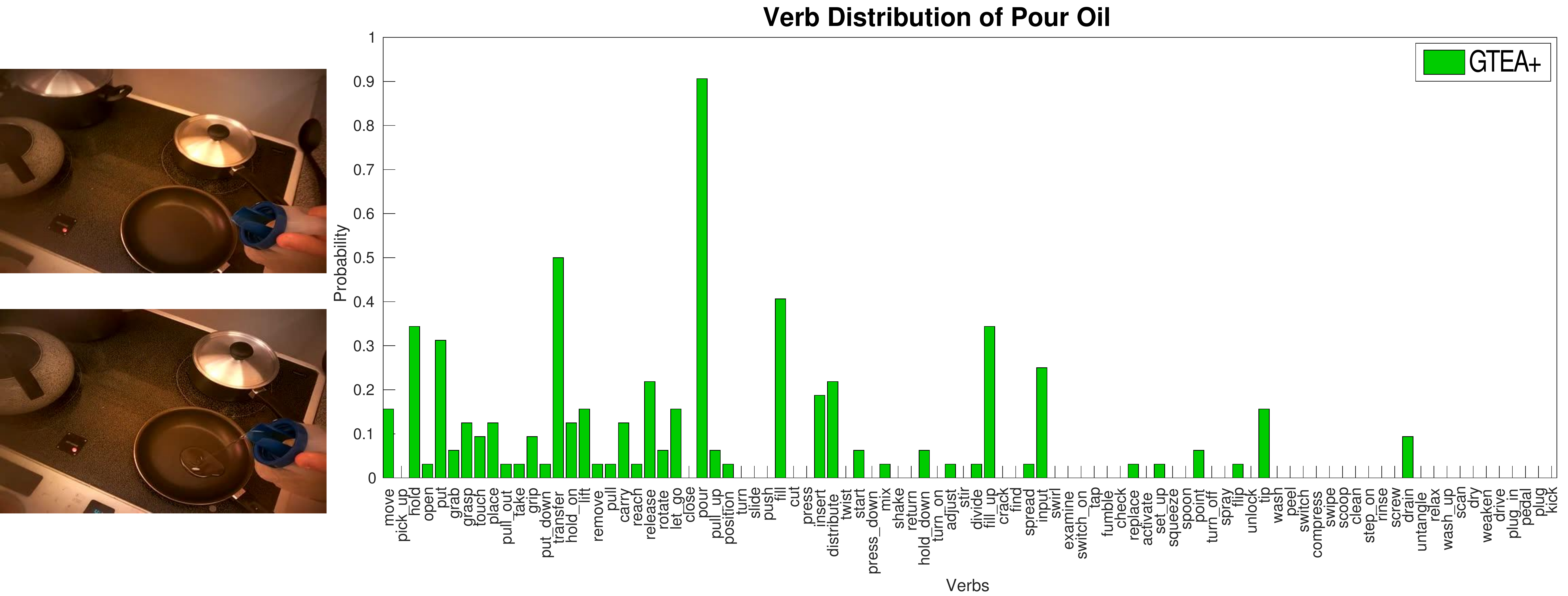}
    \label{fig:take_cup}
  \end{subfigure}
  \\[-3ex]
  \begin{subfigure}[b]{0.48\textwidth}
    \centering 
    \includegraphics[width=\textwidth]{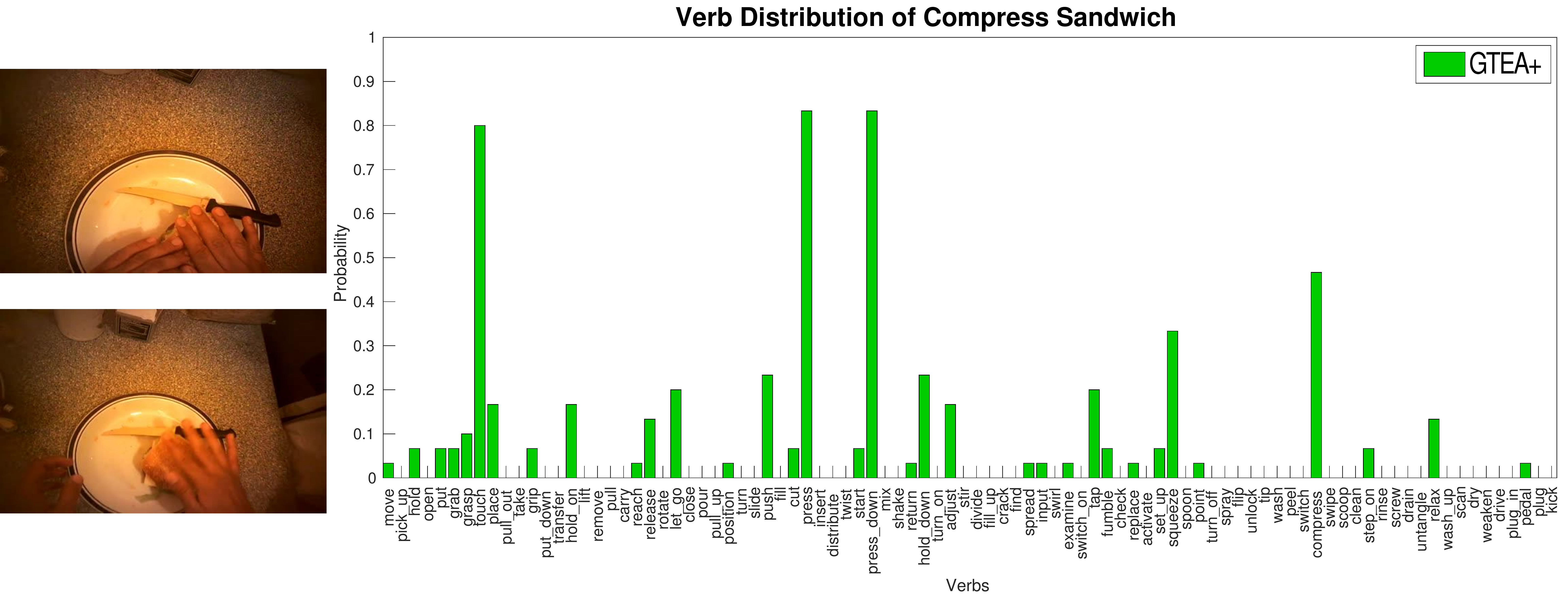}
    \label{fig:take_cup}
  \end{subfigure}
  \\[-3ex]
\caption{Annotation distributions over $90$ verbs for three object interactions, collected from AMT (30-50 annotators).}
\label{fig:verb_distribution}
\end{figure}

Accordingly, for each object interaction, the annotations record the frequency, i.e. number of times, an annotator selected the verb to label that object interaction. We normalised these frequencies by the number of annotators per video, and refer to these as \textit{the probability of annotating the video sequence with the corresponding verb by a human annotator}.
We do not filter the annotations as due to the large amount of annotators per video noisy verbs will be negligible.

In Figure~\ref{fig:verb_distribution}, examples of three distributions of probabilities over the $90$ verbs can be seen. The original class `Take Measuring Cup' from~\cite{de2008guide} is correctly annotated with many common verbs such as \textit{move}, \textit{pick~up} and \textit{take}.
\textit{Put~down} also has a high probability,
as the video concludes with the user placing the cup down on the counter.
The action `Pour Oil', from~\cite{Fathi2012}, has a high probability for verbs \textit{pour} and \textit{hold}, representing the user holding the oil container whilst performing the action.
For the action `Compress Sandwich' from~\cite{Fathi2012}, three visible labels have high probability -- \textit{press, press~down} and \textit{push}. The original label \textit{compress} was picked with a probability of 0.47.


\subsection{Annotation Statistics}
\label{subsec:ann_statistics}
In Figure~\ref{fig:annotation}(a) we show the number of annotations for each of the $90$ verbs -- an expected tail distribution -- with the top-5 used verbs being: \textit{move, pick~up, hold, open} and \textit{put}.
In Figure~\ref{fig:annotation}(b) we see that for CMU the classes which had the highest number of verbs chosen comprised of the most sub-actions.
`Crack Egg' involves the user both cracking the egg on the side as well as opening the egg shell and finally pouring the egg into the bowl.
Similarly `Spray Pam' includes the user opening the can, shaking it and spraying.
Further analysis showed little correlation between the length of the action and the number of annotations that were chosen, $R^2=0.0854$ as well as no correlation between the length of the sequence and the number of unique verbs that were chosen, $R^2=0.0972$.
It is interesting to note that the lowest number of unique verbs chosen, $20$ for `Open Microwave', is still higher than the number of verbs present in the original annotations for both CMU and GTEA+.
Figure~\ref{fig:annotation}(c) shows that all annotators chose several verbs per video, with the average number of annotations ranging from $10$ for `Take Pam' to an average of $3$ for `Close Fridge'.

We also checked whether any of the $90$ verbs chosen are redundant, \ie~could be fully replaced by another verb.
We do this by using symmetric 
co-occurrences of these verbs in the AMT annotations.
Assume $\mathcal{C}_{ij}$ is the number of times the $i$th verb was annotated with the $j$th verb across all annotators and classes for both datasets.
If $\mathcal{N} (i,j) = {\mathcal{C}_{ij}}/{\sum_k \mathcal{C}_{ik}}$, then the symmetric co-occurrence is calculated, $\mathcal{S}(i,j) = \frac{\mathcal{N}(i,j) + \mathcal{N}(j,i)}{2}$.
We report the top $40$ co-occurrences in Figure~\ref{fig:annotation}(d),
by combining co-occurrences from both datasets.
Verbs with high symmetrical occurrences should be interchangeable regardless of context, \eg~(\textit{switch~on}, \textit{turn~on}), (\textit{put}, \textit{place}) and  (\textit{rotate}, \textit{turn}).
In two cases, \textit{(wash, rinse)} and \textit{(kick, check)}, the co-occurrence originates from only one dataset, namely GTEA+. However, all the remaining top co-occurrences are present in both datasets.
Importantly, we find that none of the $90$ verbs were fully redundant or interchangeable, {$\mathcal{S}(i,j)_{CMU} + \mathcal{S}(i,j)_{GTEA+} < 2$}, with the highest co-occurrence being $1.46$.

\section{Method}
\label{sec:method}

We first review the standard recognition approach and then introduce the new probabilistic multi-label action recognition approach.
For both approaches, given a list of verbs, \;$\mathcal{C}$, we wish to learn a mapping between video sequences and the annotation verbs.
In the first case, only one annotation verb can be used as a label. In the second probabilistic approach, which we propose here, we not only allow multiple annotation verbs but wish to learn the probability of a human using that verb to annotate the video. In this section, we refer to the traditional approach using $\mathcal{T}$ and for the proposed approach using $\mathcal{P}$.

We define the standard recognition problem $\mathcal{T}$ as a set of videos, $X = \{x_i; i = 1..K\}$, each with a single label $Y = \{y_i; i = 1.. K\}$, out of a total number of classes $|\mathcal{C}|$, thus ${1 \le y_i \le |\mathcal{C}|}$.
The learnt model, $f$, with a set of parameters, $\omega$, would then aim for the prediction $\hat{y_i} = f(\omega, x_i)$, typically trained using a logistic regression loss function:
\begin{equation}
	L_\mathcal{T}(\omega) = -\frac{1}{K} \sum_i \big(y_i \log(\hat{y_i}) + (1-y_i) log (1-\hat{y_i})\big)
    \label{eq:loss}
\end{equation}

\vspace*{-8pt}
\noindent For a test set with $M$ videos, the accuracy for this recognition model, can then be calculated to be 
\begin{equation}
A_\mathcal{T}(\omega) = \frac{1}{M}\sum_i (\hat{y_i} = y_i)
\end{equation}

\vspace*{-8pt}
We can similarly define the \textit{same standard recognition approach} using a one-hot vector; labels $\bm{y_i}$ of length $|\mathcal{C}|$.
The $j$th element in $\bm{y_i}(j)$ is $1$ if the single label of the video is the $j$th class in $\mathcal{C}$ which we denote as $\mathcal{C}_j$ for brevity.
The recognition model can then predict a vector $\bm{\hat{y_i}}$ with continuous elements limited to the range $[0,1]$.
The accuracy is updated to account for the label vectors,
\begin{equation}
A_\mathcal{T}(\omega) = \frac{1}{M} \sum_i (\arg\max{\bm{\hat{y_i}}} = \arg\max{\bm{y_i}})
\label{eq:accuracy1}
\end{equation}

\vspace*{-8pt}
\noindent Note that we use $\arg\max$ here in that is common that a classifier may output a score of $x_i$ belonging to a certain class and as such we choose to classify the video as the most likely class.

As semantically ambiguous verbs are added, and so the number of verbs $|\mathcal{C}|$ increases, verbs that are semantically correlated (\eg~\textit{put}, \textit{place}) or are only correlated in certain contexts (\eg~\textit{turn~on}, \textit{rotate}) will result in an increased error in classification.
To accommodate for a large number of verbs and overlaps in annotations (ref. Figure~\ref{fig:magic_example}), we re-define $\bm{y_i}$ not as a one-hot vector but instead as a distribution of probabilities where $\bm{y_i}(j) = P(\mathcal{C}_j|x_i)$.
The probabilities assigned to the class labels allow for capturing verbs that are less frequently used to describe the action albeit still correct.
Note that while $0 \le \bm{y_i}(j) \le 1$, the distribution $\bm{y_i}$ is not normalised, i.e. $\sum_j \bm{y_i}(j) \neq 1$.
This is intentional so correct multiple labels are not penalised.
Each element $\bm{y_i}(j)$ measures the probability of a human annotator selecting that verb, and accordingly, {$\forall \mathcal{C}_j, P(\lnot \mathcal{C}_j | x_i) = 1 - P(\mathcal{C}_j | x_i)$}.

In this case the output $\bm{y_i}$ is a set of linear neurons, trained from the frequency of annotations. We adjust the loss function to calculate the Euclidean loss between the two distributions of probabilities $\bm{y_i}$ and $\bm{\hat{y_i}}$,

\begin{equation}
L_\mathcal{P}(\omega) = -\frac{1}{K} \sum_i \bigl((\bm{y_i}-\bm{\hat{y_i}})^T(\bm{y_i}-\bm{\hat{y_i}})\bigr)^{\frac{1}{2}}
\label{eq:loss_euclidean}
\end{equation}

In Eq.~\ref{eq:loss_euclidean}, errors in estimating probabilities are penalised without a preference for high or low probabilities.


We evaluate the proposed probabilistic classifier in two ways, first using a threshold on probabilities $\alpha$. Assume,
\begin{equation}
\bm{Y}_{i}^\alpha = \{\mathcal{C}_j; \quad \forall \bm{y_i}(j) > \alpha\}
\end{equation}
is the list of all verbs annotated with a probability higher than $\alpha$, and the corresponding top estimated list of verbs,
\begin{equation}
\bm{\hat{Y}}_{i}^\alpha = \{\mathcal{C}_j; \quad \forall \bm{\hat{y}_i}(j) \in top_k(\bm{\hat{y}_i}) \land k = |\bm{Y}_{i}^\alpha|\}
\end{equation}
The accuracy of the proposed system can then be calculated for a certain threshold $\alpha$ as
\begin{equation}
	A_\mathcal{P}(\omega, \alpha) = \frac{1}{M} \sum_i \frac{|\bm{Y}_{i}^\alpha \cap \bm{\hat{Y}}_{i}^\alpha|}{|\bm{Y}_{i}^\alpha|}
    \label{eq:accuracyOverall}
\end{equation}

Specifically, assume $k$ verbs were annotated with a probability greater than a threshold $\alpha$, we then find the $top_k$ verbs estimated by the model and measure the overlap.
For example, assume $\alpha = 0.7$ and a video is annotated with three verbs where $\bm{y}_i(j) > \alpha$, namely: \textit{put, place} and \textit{move}. We find the top 3 estimated verbs and compare the sets. If the three top estimated verbs were \textit{put, move} and \textit{open}, the accuracy for this video would be set to $\frac{2}{3}$. Eq.~\ref{eq:accuracyOverall} measures the ability of the model to retrieve the right set of annotation verbs.
We also measure the mean squared error in predicting probabilities for each verb using, 
\begin{equation}
E_\mathcal{P} (C_j) = \frac{1}{M} \sum_i ||\bm{y}_i(j) - \bm{\hat{y}}_i(j)||
\label{eq:mse}
\end{equation}

We next present results for the probabilistic classifier, starting with the datasets and the experimental setup.
\section{Experiments and Results}
\label{sec:experiments}

For our experiments we use the two publicly available egocentric datasets aforementioned CMU~\cite{de2008guide} ($404$ video segments) and GTEA+~\cite{Fathi2012} ($1001$ video segments) with the AMT annotations described in Section~\ref{sec:annotations}. 
Both datasets include videos of daily activities in the kitchen, recorded using a head mounted camera.
For each dataset, $5$ cross-fold validation was used where we equally distribute videos from each class across the folds.
In Section~\ref{subsec:implementation}, we describe the implementation details for the two stream CNN model used, and then report subsections on results, namely: testing against classification (\ref{subsec:exp1}), evaluation of estimated probabilities (\ref{subsec:exp3}) and evaluation of verb co-occurrences (\ref{subsec:exp2}).

\begin{figure*}[t]
\begin{center}
   \includegraphics[width=1.0\linewidth]{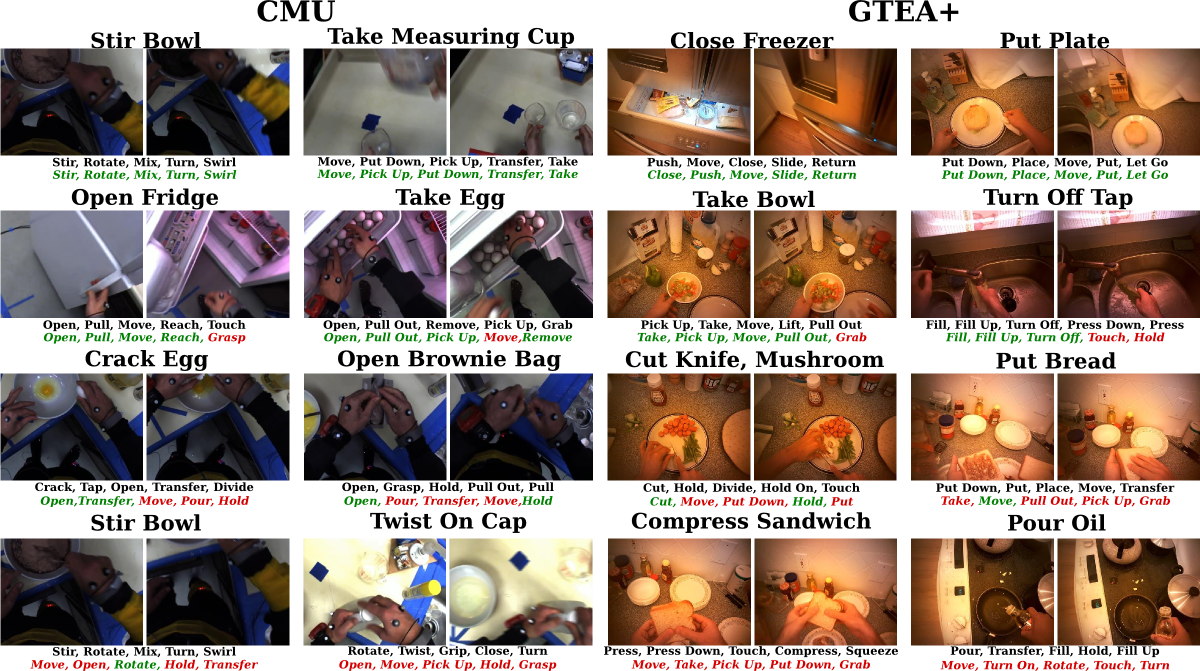}
\end{center}
   \caption{Qualitative results of the proposed method. 
   Top $5$ ground truth annotations (in black) from the AMT workers. The top 5 predicted verbs from the method are shown in green if correct and red if incorrect.}
\label{fig:qualitative}
\end{figure*}

\subsection{Implementation Details}
\label{subsec:implementation}
We trained and tested using the two stream fusion CNN model from~\cite{feichtenhofer2016convolutional} that uses two VGG-16 architecture networks as a base, pre-trained on the UCF101 dataset~\cite{soomro2012ucf101}.
The last layer of the network was replaced with a Euclidean loss layer, and a number of nodes equal to the number of annotation verbs $|\mathcal{C}| = 90$. The base spatial and temporal networks were each fine-tuned on the videos in the training splits before being used to train the fusion network. 
Each network was trained for 10 epochs which we found was enough for the network to converge due to using UCF101 pre-trained models.
We set a fixed learning rate for spatial, $10^{-3}$, and used a variable learning rate for temporal and fusion networks in the range of $10^{-({2,3,4,5})}$ to $10^{-({3,4,5,6})}$ depending on the training epoch. For the spatial network a dropout ratio of 0.85 was used for the first two fully-connected layers.
We fused the temporal network into the spatial network after ReLU5 using convolution for the spatial fusion as well as using both 3D convolution and 3D pooling for the spatio-temporal fusion and we propagated back to the fusion layer.
For training we use a batch size of 128, a weight decay of $0.0005$, a momentum of 0.9 and did not use batch normalisation as in~\cite{feichtenhofer2016convolutional}.


\subsection{Testing Against Classification}
\label{subsec:exp1}
We first test our method against a standard classification approach.
For this baseline, we select the majority vote per video, that is, the verb with the highest number of annotators, and construct a one-hot vector $\bm{y_i}$ for training labels.
This matches the standard classification approach where each video is assigned a single label.
The accuracy is calculated for traditional classification using Eq.~\ref{eq:accuracy1}, and for the proposed probabilistic classifier using Eq.~\ref{eq:accuracyOverall}.
We report results for $\alpha = 0.5$ as this is the threshold where, for both datasets, each video has at least one annotated verb.
predicting multiple labels.
At $\alpha = 0.5$, the number of verbs per video is
$\mu_{CMU}=3.49$, $\sigma_{CMU}=1.66$ and $\mu_{GTEA+}=3.14$ and $\sigma_{GTEA+}=1.41$.
\begin{table}[t]
\centering
\begin{tabular}{|l|c|c|}
\hline
         & CMU  & GTEA+ \\ \hline
         Prev. &48.6 \cite{spriggs2009temporal} &60.5 \cite{li2015delving}, 65.1~\cite{Ma16}\\ \hline
Classification ($\mathcal{T}$) & 52.0 & 61.8  \\ \hline
Proposed ($\mathcal{P}$) & 63.3 & 67.9  \\ \hline
\end{tabular}
\caption{Accuracy results comparing the proposed method to standard classification.
Results of the proposed use Eq.~\ref{eq:accuracyOverall} [$\alpha = 0.5$].}
\label{tab:baseline}
\end{table}

Table~\ref{tab:baseline} 
shows that our classification results are comparable to published results on these datasets, albeit using the majority verb from AMT annotations.
It also shows that the proposed method is able to obtain a higher accuracy over the classification approach while 

In viewing the result, note that the proposed method is attempting a harder problem than the classification task. Assume that $4$ verbs for one video were annotated with a probability $\bm{y}_i(j) > \alpha$, correctly retrieving the majority vote would add $1.0$ to the accuracy of the classifier, but only $0.25$ for the proposed method. The proposed method would achieve an accuracy of 1.0 for the video if the top 4 estimated verbs matched the top 4 annotated verbs.
There is no correlation between the number of verbs chosen above the threshold $\alpha$ and the accuracy, for CMU or GTEA+ ($R^2=0.144$ and $R^2=0.0004$ respectively).

\begin{table*}[t]
\centering
\resizebox{\textwidth}{!}{%
\begin{tabular}{|r|c|c|c|c|c|c|c|c|c|c|c|c|c|c|c|c|c|c|}
\hline
 & \multicolumn{9}{c|}{CMU} & \multicolumn{9}{c|}{GTEA+} \\ \hline
$\alpha$ & 0.1 & 0.2 & 0.3 & 0.4 & 0.5 & 0.6 & 0.7 & 0.8 & 0.9 & 0.1 & 0.2 & 0.3 & 0.4 & 0.5 & 0.6 & 0.7 & 0.8 & 0.9 \\ \hline
Number of Videos & 404 & 404 & 404 & 404 & 404 & 390 & 304 & 269 & 53 & 1001 & 1001 & 1001 & 1001 & 1001 & 975 & 748 & 732 & 319 \\ \hline
Avg. Verbs per Video & 20.1 & 13.0 & 8.58 & 5.80 & 3.49 & 2.44 & 1.73 & 1.04 & 1 & 13.2 & 9.08 & 6.32 & 4.46 & 3.14 & 1.92 & 1.90 & 1.22 & 1 \\ \hhline{|=|=|=|=|=|=|=|=|=|=|=|=|=|=|=|=|=|=|=|}
Scores from Classification & 31.2 & 29.1 & 31.6 & 35.1 & 33.9 & 36.0 & 50.3 & \textbf{69.3} & \textbf{54.7} & 26.2 & 25.4 & 29.1 & 33.5 & 40.3 & 53.1 & \textbf{54.6} & \textbf{65.7} & \textbf{78.7} \\ \hline
Proposed Method & \textbf{79.1} & \textbf{73.7} & \textbf{69.1} & \textbf{66.2} & \textbf{63.3} & \textbf{57.4} & \textbf{63.7} & \textbf{69.3} & \textbf{54.7} & \textbf{74.8} & \textbf{71.9} & \textbf{70.4} & \textbf{68.3} & \textbf{67.9} & \textbf{59.2} & 52.5 & 59.8 & 71.2 \\ \hline

\end{tabular}
}
\caption{Accuracy results of the proposed method against scores from training one-hot vector baseline with $0.1 \le \alpha \le 0.9$.}
\label{tab:baseline_full}
\end{table*}

We next show qualitative results for the top $5$ verbs in Figure~\ref{fig:qualitative}.
The proposed method is able to predict the correct verbs in the correct order.
We find that it often predicts \textit{move} highly for actions it cannot recognise.
In several examples, we note that for cases where the estimated verb does not match the top $5$ ground truth annotations, it still is present with high probability in annotations, typically within the top $10$ verbs, for example \textit{pour} in `Crack Egg', \textit{grasp} in `Open Fridge' and \textit{grab} in `Take Bowl'.

While Table~\ref{tab:baseline} compares the proposed method against classification, we also test whether the model trained for classification could be used to estimate probabilities. Instead of capturing $\arg\max\bm{\hat{y}_i}$, we consider the scores from the network $\bm{\hat{y}_i}$ and compare the output using Eq.~\ref{eq:accuracyOverall} for $\alpha \in[0.1,0.9]$ in
Table~\ref{tab:baseline_full}.  
Note that the number of videos with at least one verb annotated with a probability $> \alpha$ differs.
For $\alpha > 0.7$, the number of videos with at least one verb annotated drops to 75\% for both datasets. When $\alpha > 0.7$, for videos with clear (few) high peaks, the classifier performs equally or better than the proposed method. This is expected as the majority vote classifier is tuned to recognise annotations with high probabilities solely.



\subsection{Evaluation of the Proposed Verb Probabilities}
\label{subsec:exp3}
\begin{figure}
\centering
  \begin{subfigure}[b]{0.45\textwidth}
    \centering
    \includegraphics[width=\textwidth]{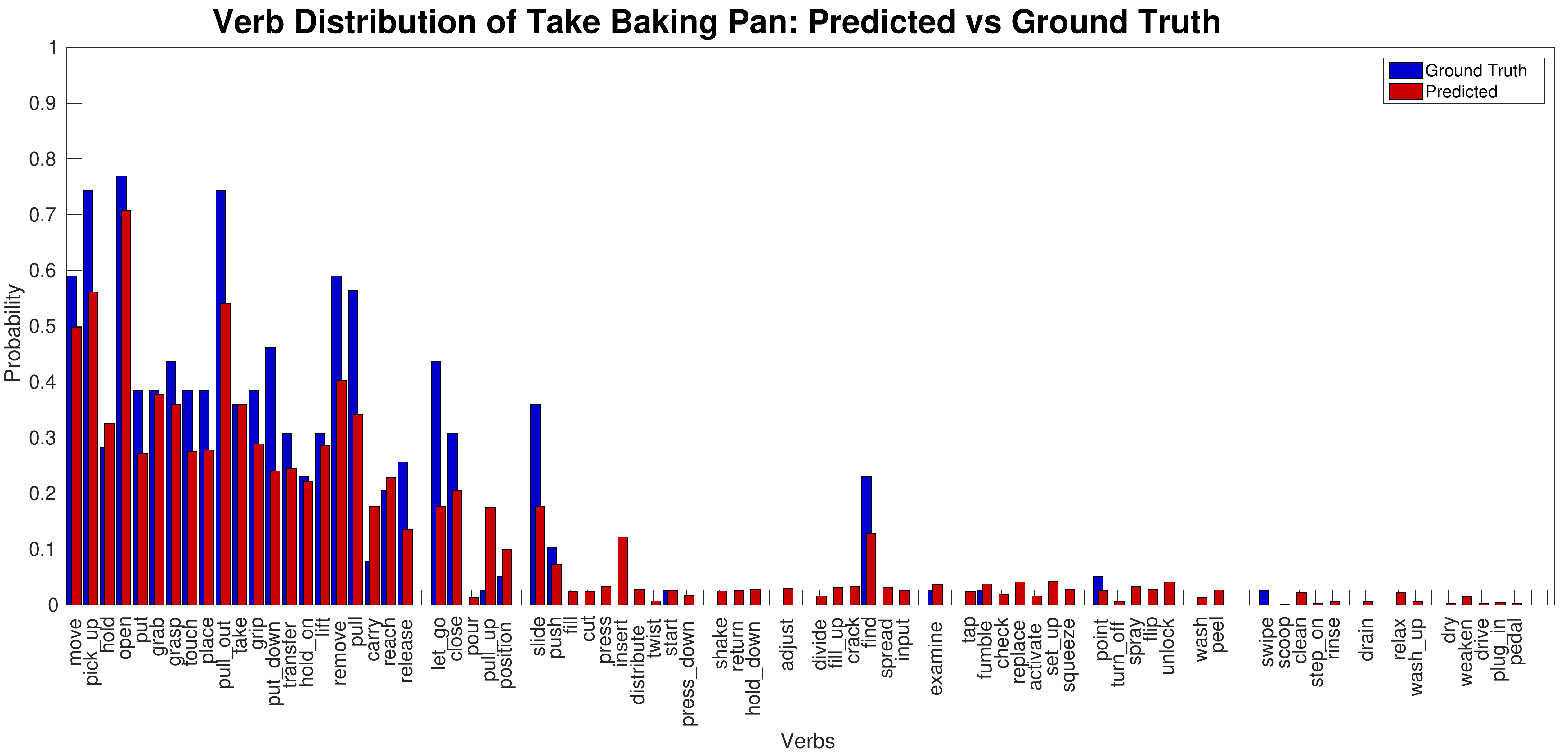}
    \label{fig:dist_stir}
  \end{subfigure}
  \\[-3ex]
  \begin{subfigure}[b]{0.45\textwidth}
    \centering 
    \includegraphics[width=\textwidth]{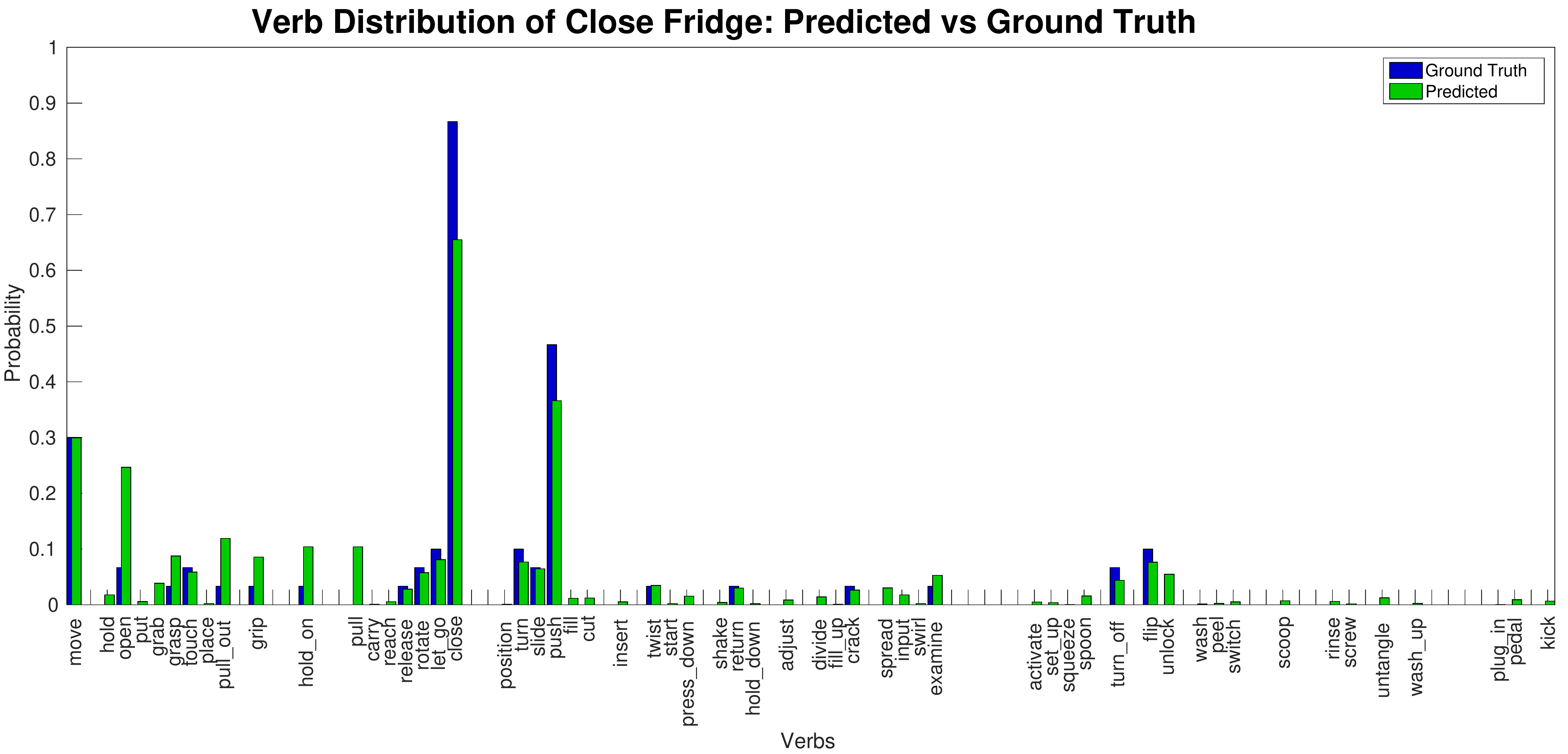} 
    \label{fig:dist_take}
  \end{subfigure}
  \\[-3ex]
  \caption{Two different videos with the ground truth and predicted verb probabilities. Top: Take baking pan from CMU. Bottom: Close fridge from GTEA+.} 
  \label{fig:distributions_comp}
\end{figure}

We next evaluate the ability of the method to retrieve the distribution of probabilities per video. 
Figure~\ref{fig:distributions_comp} includes two example distributions of probabilities where the method has been able to fit well to the ground truth. In the top example, the method is tested on a video with a large number of annotated verbs, while the bottom example shows that the method is equally successful in recognising distributions with few visible peaks.
Generally, the method seems to underestimate the probabilities of the dominant verbs, causing slight differences in the predicted verb rankings.

\begin{figure}[t]
\begin{center}
   \includegraphics[width=\linewidth]{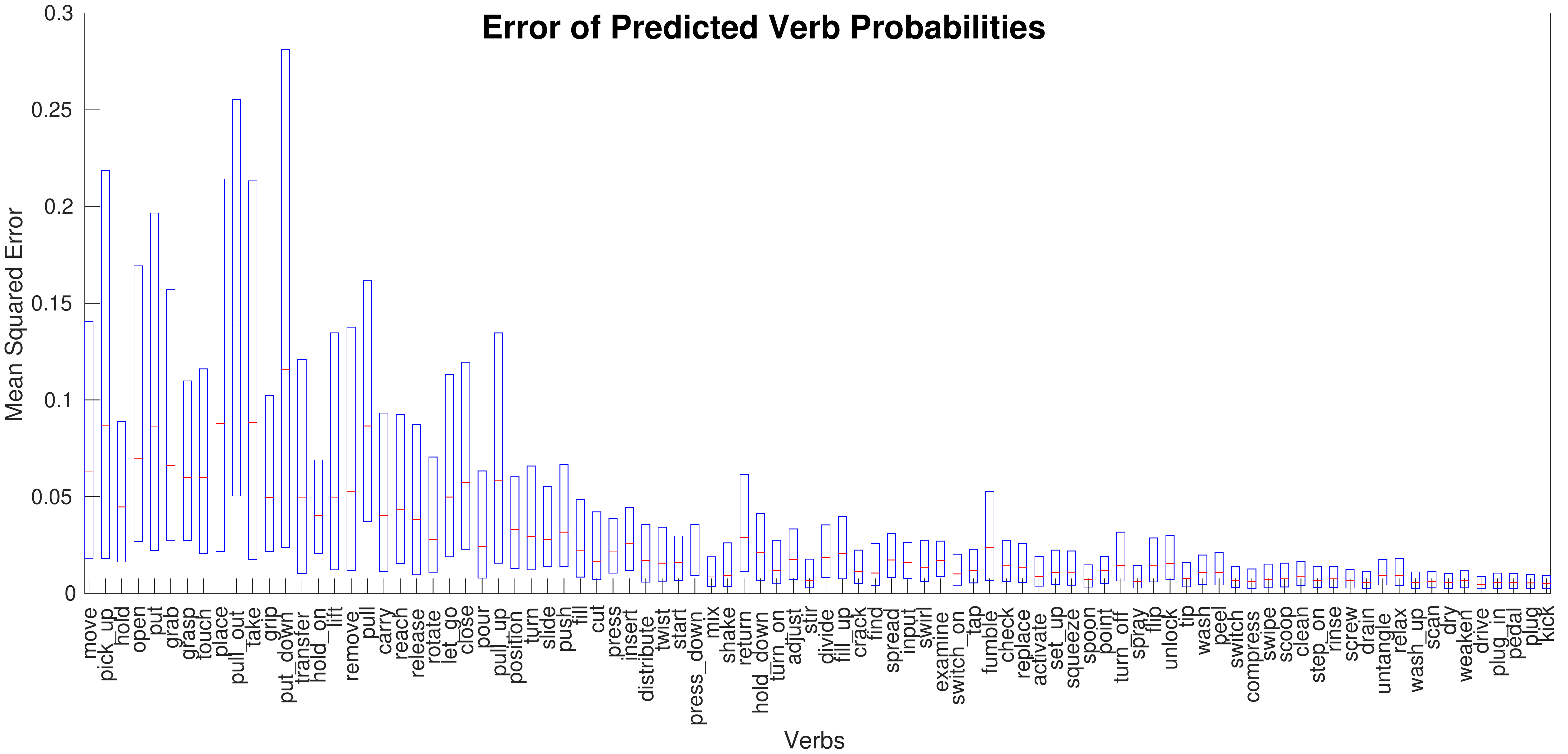}
\end{center}
   \caption{Mean square error of each of the predicted verbs calculated using Eq.~\ref{eq:mse}.}
\label{fig:mse}
\end{figure}

\begin{figure*}[t]
\begin{center}
   \includegraphics[width=0.86\linewidth]{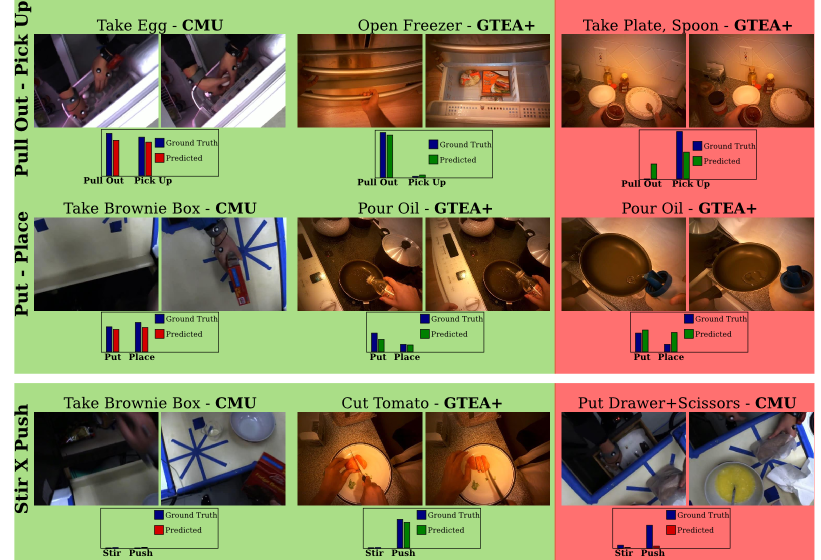}
\end{center}
   \caption{Examples of the method predicting co-occurrences between verbs. The top two rows show pairs of verbs with high co-occurrence with the last row showing an example of two verbs with low co-occurrence. Ground truth and predicted probabilities are also shown for each verb. The proposed method is able to not only recognise when two verbs should co-occur with high probability, first column, but also predict sensible probabilities when one of the verbs has a lower probability, second column. Failure cases are shown in the last column.}
\label{fig:qualitative_co_occur}
\end{figure*}

Figure~\ref{fig:mse} shows the mean square error for the predicted probabilities of each verb (Eq.~\ref{eq:mse}).
The median MSE is below $0.1$ for $88$ verbs - excluding \textit{put~down} and \textit{pull~out} where the median is below $0.15$. 
We note that the error in predicting the probability for the verb \textit{move} is low, compared to the frequency of the annotation.
Results also show low error bars for specific actions such as \textit{hold}, \textit{grip} and \textit{open} suggesting that the proposed method is able to learn these actions well.
Verbs with the highest error, \textit{pick~up}, \textit{pull~out} and \textit{place} for example, correspond to actions that often occur at the start or end of other actions; \textit{pick~up} a knife before \textit{cutting}.

We finally test whether the method predicts annotations with high or low probabilities more accurately, and note no correlation between the annotated probability of a verb with $\alpha>0.1$ and the MSE ($R^2=0.0383$) for all verbs and all videos.

\subsection{Evaluation of Learning Co-Occurrences}
\label{subsec:exp2}

\begin{figure}[t]
\begin{center}
   \includegraphics[width=\linewidth]{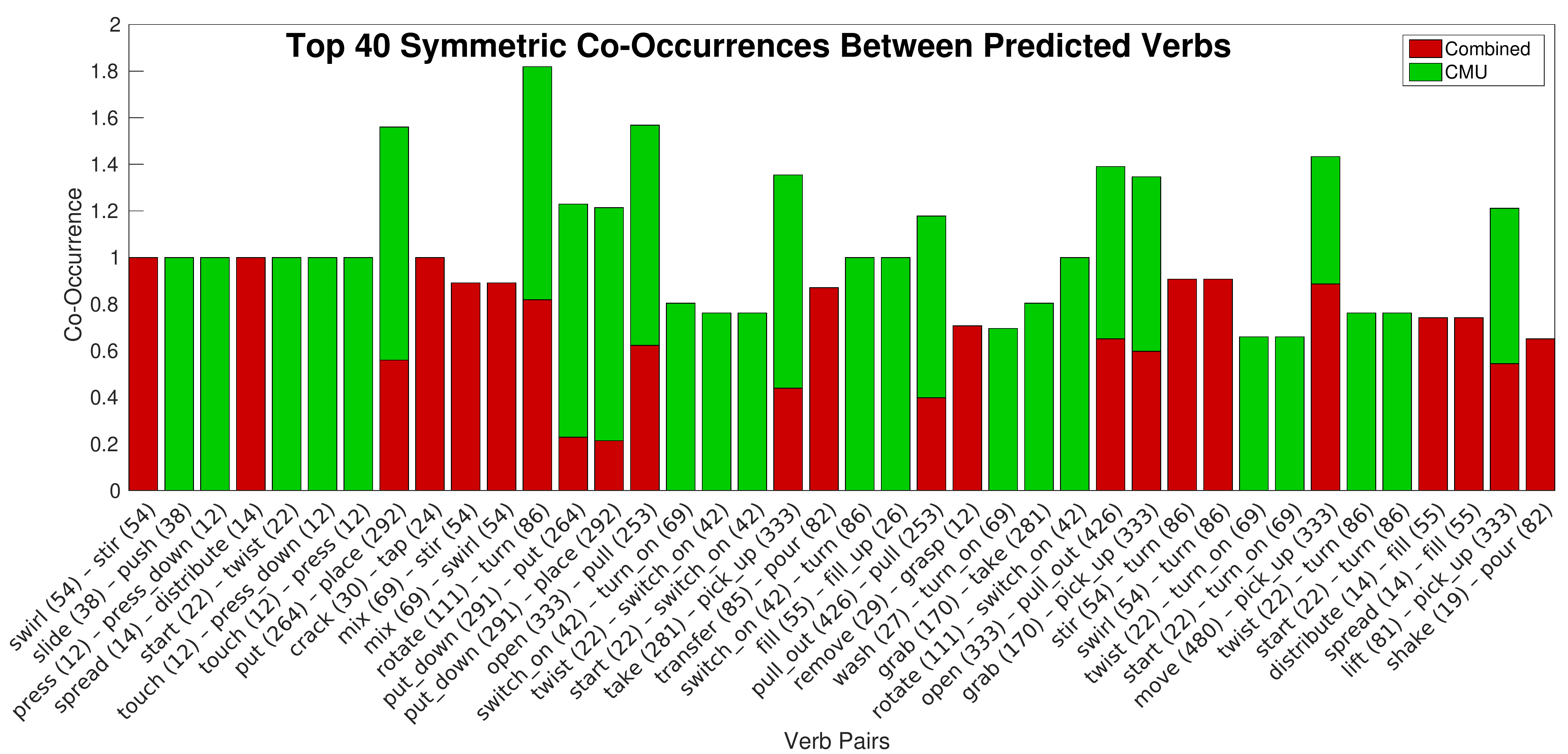}
\end{center}
   \caption{The top 40 symmetric co-occurrences predicted from the method found with $\alpha=0.5$ for both CMU and GTEA+. The verbs are ordered by both datasets but plotted to show the contribution of each dataset separately.}
\label{fig:co_occur_pred}
\end{figure}

We next look at class overlaps, which correspond to predicting annotation co-occurrences. A co-occurrence between two verbs $\mathcal{C}_{ij}$ occurs when a video is annotated using both verbs with a probability $> \alpha$.

Figure~\ref{fig:qualitative_co_occur} shows examples of correct and incorrect predicted co-occurrences.
For the two verbs (\textit{pull~out}, \textit{pick~up}), we study three cases of their co-occurrence. For the action `Take Egg', the person pulls the egg out of the box and picks it up. Ground truth annotations indicate high probability for both verbs. Our proposed method is indeed able to predict the co-occurrence with high accuracy. For a different interaction of `Opening the Freezer's drawer', the method correctly predicts a high accuracy for annotating the video using the verb \textit{pull~out}, and correctly estimates that the verb \textit{pick~up} is not a suitable verb to annotate this video. The figure shows cases of failure where the co-occurrences are incorrectly predicted.
The final row shows two verbs, \textit{stir} and \textit{push}, that have a very low co-occurrence.
The model is able to define the soft boundaries between these classes. 

Using the same co-occurrence metric in Figure~\ref{fig:annotation}(d), we study the top co-occurrences that the method has learned. Figure~\ref{fig:co_occur_pred} shows the top 40 symmetric co-occurrences with $\alpha=0.5$ for both datasets.
We find that a few verbs with a small number of instances are always (and sensibly) predicted together, such as (\textit{swirl}, \textit{stir}) and (\textit{spread}, \textit{distribute}).
This suggests 
that the method is able to learn meaningful semantic relationships between verbs.
As opposed to Figure~\ref{fig:annotation}(d), where co-occurrences are detected in annotations from both datasets, Figure~\ref{fig:co_occur_pred} shows co-occurrences are strongly discovered in one of the two datasets. Recall that we fine-tune different models for the two datasets. Training a single model might result in common co-occurrences. 

While in this work we attempted multi-label classification, the model does not localise the various labels in space or time. This will be particularly interesting for cases where antonyms are picked with high probability - picking up a baking pan before putting it down. This direction of investigation is left for future work.

\section{Conclusion and Future Work}
\label{sec:conc}
\vspace*{-6pt}
In this paper, we present the case for multi-label annotations of object interactions, in egocentric video, and provide a comprehensive set of AMT multi-verb annotations for two datasets.
We provide a model that estimates the probability of a label being used to annotate the video, out of a list of possible verbs.
When using a two-stream fusion CNN, we show that the method is able to predict not only the correct verb labels but also sensible probabilities for two public datasets outperforming single label classification.
We discuss success and failure cases for multi-label classification, and show the method is able to learn co-occurrences of semantically related verbs and simultaneous interactions.

In the introduction, we highlighted three incorrect assumptions when using single-labels to annotate object interactions. We plan to investigate next how the proposed probabilistic multi-label classifier addresses each of these assumptions, towards further refinement of the model and its performance.

{\small
\bibliographystyle{ieee}
\bibliography{iccv_paper}
}

\end{document}